\documentclass{article}

\usepackage[nonatbib,preprint]{neurips_2021} 

\usepackage[utf8]{inputenc} % allow utf-8 input
\usepackage[T1]{fontenc}    % use 8-bit T1 fonts
\usepackage{authblk}
\usepackage{hyperref}       % hyperlinks 
\usepackage{url}            % simple URL typesetting
\usepackage{booktabs}       % professional-quality tables
\usepackage{amsfonts}       % blackboard math symbols
\usepackage{nicefrac}       % compact symbols for 1/2, etc.
\usepackage{microtype}      % microtypography
\usepackage{xcolor}         % colors
  
\usepackage{graphicx} 
\usepackage{float} 
\usepackage{subfigure} 
\usepackage{multirow}
\usepackage{array}
\usepackage{amsmath}
\usepackage{bm}
\usepackage[linesnumbered,ruled,vlined]{algorithm2e}

\title{Deep Contrastive Graph Representation via Adaptive Homotopy Learning}

\author[a]{Rui Zhang}
\author[a]{Chengjun Lu}
\author[a]{Ziheng Jiao}
\author[a]{Xuelong Li \thanks{Corresponding author: Li@nwpu.edu.cn}}
\affil[a]{Department of Computer Science, Northwestern Polytechnical University}
\begin{document}
\maketitle
\begin{abstract}

      Homotopy model is an excellent tool exploited by diverse research works in the field of machine learning. However, its flexibility is limited due to lack of adaptiveness, i.e., manual fixing or tuning the appropriate homotopy coefficients. To address the problem above, we propose a novel adaptive homotopy framework (AH) in which the Maclaurin duality is employed, such that the homotopy parameters can be adaptively obtained. Accordingly, the proposed AH can be widely utilized to enhance the homotopy-based algorithm. In particular, in this paper, we apply AH to contrastive learning (AHCL) such that it can be effectively transferred from weak-supervised learning (given label priori) to unsupervised learning, where soft labels of contrastive learning are directly and adaptively learned. Accordingly, AHCL has the adaptive ability to extract deep features without any sort of prior information. Consequently, the affinity matrix formulated by the related adaptive labels can be constructed as the deep Laplacian graph that incorporates the topology of deep representations for the inputs. Eventually, extensive experiments on benchmark datasets validate the superiority of our method.

\end{abstract}

\section{Introduction}
As an important optimization method in the field of machine learning, the homotopy method is a general problem-independent technique for solving non-convex problems. Additionally, it has been widely applied in representation learning due to the excellent optimization capabilities \cite{chow1978finding,nocedal2006numerical}. However, due to relying heavily on the priori homotopy coefficient, most homotopy-based algorithms are lack of adaptability, and even cannot  be directly applied to unsupervised learning.\cite{mobahi2015link}. Meanwhile, unsupervised representation learning seems more promising compared to the high cost and limitations of supervised learning in practical applications \cite{hofmann2001unsupervised,barlow1989unsupervised,le2013building,arora2019theoretical}. To make the homotopy algorithm perform well in unsupervised learning as well, it is significant to directly endow homotopy algorithm with adaptive ability.

Particularly, contrastive learning is an efficient homotopy-based method for representation learning. The basic idea of contrastive learning is to project the original data into a feature space in which positive pairs have the greatest similarity and negative pairs have the least \cite{hadsell2006dimensionality}. There are a lot of well known researches in the field of unsupervised comparative learning \cite{arora2019theoretical,henaff2020data}. However, it cannot directly realize unsupervised learning due to lack of adaptiveness.

To address the issue concerning the adaptiveness, in this paper, we propose a novel adaptive homotopy framework (AH) in which the Maclaurin duality is employed. The main contributions are listed as follows:

$\bullet\ $ To endow the homotopy model with adaptability directly, the AH model employs the Maclaurin duality to learn the homotopy parameter adaptively, and then the homotopy model will no longer be limited to priori weight information unlike before.

$\bullet\ $ The proposed AH model can extend the various existing homotopy-based algorithms such that they as well have the adaptive ability of unsupervised learning. In particular, we apply AH to contrastive learning (AHCL) such that it can be effectively transferred from weak-supervised learning (given label prior) to unsupervised learning. Accordingly, the soft labels can be upgraded to directly learnable adaptive weight. Thus, AHCL has the adaptive ability to extract embedding in latent space without supporting of prior information.

$\bullet\ $ Our AHCL is able to learn the deep graph representation among nodes of the inputs. It makes the adaptive soft label correspond to the affinity relationship of inputs. Consequently, the affinity matrix formulated by them can be constructed as the deep Laplacian graph that incorporates the topology of deep representations. Furthermore, we eliminate the collapse caused by the adaptive weight via constructing a decoder to reconstruct inputs from learned embedding.

\begin{figure}[t]
      \centering
      % \fbox{\rule[-.5cm]{0cm}{4cm} \rule[-.5cm]{4cm}{0cm}}
      \includegraphics[width=1\textwidth]{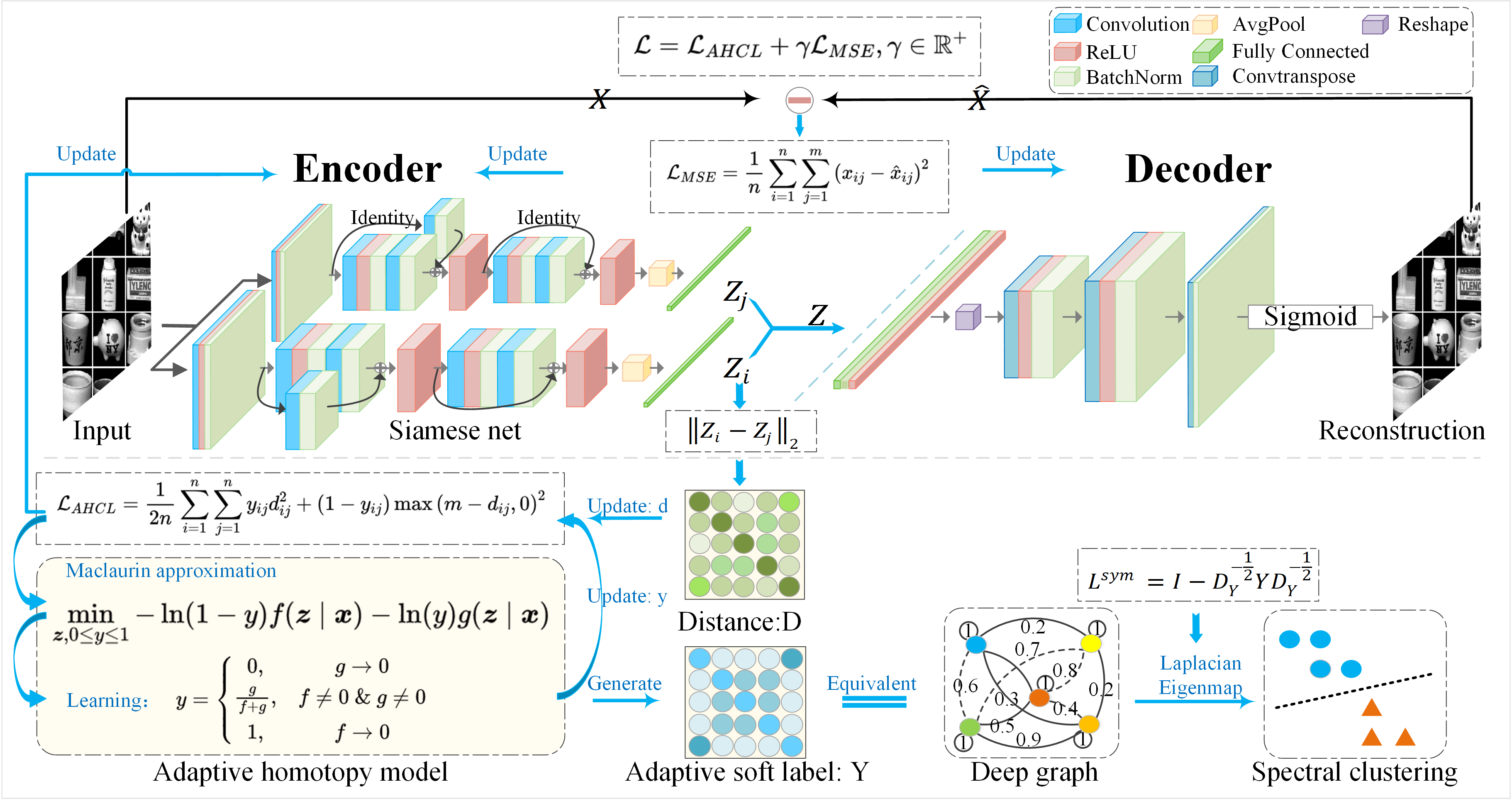}
      \caption{The Framework of Adaptive Homotopy-Contrastive Learning (AHCL). We extract embedding $\bm{Z}$ of data pairs using a Siamese network as an Encoder consisted of some ResNet blocks. And an unbalanced autoencoder is used as the decoder, which reconstructs input data $\bm{X}$ as $\bm{\hat{X}}$. Particularly, the deep embedding leaned form AHCL can generate deep graph representation between input nodes adaptively.}
      \label{Fig:1}
\end{figure}

To verify the performance of our model, we design a network composed of an encoder and a decoder like Variational Auto Encoder. AHCL is utilized as the main objective function implemented by Mean Squared Error (MSE). The specific network architecture flowchart is illustrated in Figure \ref{Fig:1}.

\section{Background and Preliminaries}
\label{gen_inst}

\subsection{Notations}
In this paper, matrices and vectors are represented in uppercase and lowercase letters respectively, such as a $m$-dimension vector like $\bm{x}_{i}$ and a matrix like $\bm{X}_{n \times m} =\left[\bm{x}_{1}, \bm{x}_{2}, \bm{x}_{3}, \ldots, \bm{x}_{n}\right]^{T} \in \mathbb{R}^{n \times m} $, shorthand for $\bm{X}$.  Vectors whose all elements equal 1 are expressed as $\bm{I}$. Moreover, the amount of data points and feature are represented as $n$ and $m$ respectively. Moreover, $\| \bm{x}\Vert_{p} $ is the $\ell_p$ norm of the vector $\bm{x}$.

\subsection{Homotopy algorithm}
The homotopy algorithm is usually used to solve difficult optimization problems. It starts from an effectively optimized objective function, and then transforms the problem into a simple form with desired properties. Finally, the problem is gradually transformed into the original form while tracing the solution path \cite{he1999homotopy}\cite{watson1989modern}.
It is assumed that there are two continuous functions $f(\bm{z})$ and $g(\bm{z})$ on the feature space $\bm{Z}$. If $f(\bm{z})$ represents that the expectation is scenario $\mathcal{S} $, and $g(\bm{z})$ means that the expectation is $\overline{\mathcal{S} }$, meaning not scenario $\mathcal{S} $, then it is clear that $f(\bm{z})$ and $g(\bm{z})$ belong to the same target space $\bm{D} $, which serves as a topological space from the feature space $\bm{Z}$. Thus, there exists a continuous mapping $\mathrm{H}$, expressed by Eq. (\ref{eq:H0}) as
\begin{equation}
      \mathrm{H}: \bm{Z} \times[0,1] \rightarrow \bm{D}
      \label{eq:H0}
\end{equation}
defined on the product space of the feature space $Z$ with the unit interval [0,1] to $\bm{D}$ , such that Eq. (\ref{eq:H1}).
\begin{equation}
      \begin{array}{l}
            \forall \bm{z} \in \bm{Z}, \mathrm{H}(\bm{z}, 1)=f(\bm{z}) \\
            \specialrule{0em}{0.5ex}{0.5ex}
            \forall \bm{z} \in \bm{Z}, \mathrm{H}(\bm{z}, 0)=g(\bm{z})
      \end{array}
      \label{eq:H1}
\end{equation}
Hence, the mapping $\mathrm{H}$ is a homotopy between the objective functions $f(\bm{z})$ and $g(\bm{z})$ with respect to the feature space $\bm{Z}$ to the objective space $\bm{D}$. Formally, $\mathrm{H}$ could be expressed as Eq. (\ref{eq:H1}), where the coefficient $y$ is a linear weight \cite{chow1978finding}\cite{watson1996hompack90}.
\begin{equation}
      \mathrm{H}: y  f(\bm{z})+(1-y)  g(\bm{z}),\  y \in[0,1]
      \label{eq:H2}
\end{equation}
At present, it is often used to optimize objective functions so that it can extract deep features more efficiently and speed up the convergence rate of the model \cite{yousefzadeh2020deep}. And homotopy-based algorithm plays a key role in the field of computer vision and shows excellent generalization ability and remarkable learning performance \cite{chow1978finding}\cite{watson1996hompack90}\cite{allgower2012numerical}.
Examples of deep learning models using homotopic training include the work of Mobahi(2016)\cite{mobahi2016training} in training cyclic neural networks and the work of Chen and Hao \cite{chen2019homotopy} in training fully connected networks \cite{mobahi2015link, mobahi2015theoretical}.

\subsection{Contrastive learning}

Contrastive methods learn representations by contrasting positive and negative samples. In previous works, the positive and negative pairs of contrastive learning need to be calibrated manually in the form of soft labels in advance, which limits the early contrastive learning to the supervised representation learning. Nevertheless, it has still yielded many fruitful works through the classical Siamese network, which has been applied to a variety of application scenarios with outstanding performance (e.g. object recognition \cite{fu2021siamese}, target tracking \cite{bertinetto2016fully, guo2017learning, zhu2018distractor}, similarity discrimination \cite{wang2020deep}).
Recently, various studies have shown that large amounts of data are crucial for the performance of the contrastive model \cite{he2020momentum}.
However, tagged data often need to spend a lot of manpower and material resources to obtain in the real scene, which lets it illusory seriously for applied in real application scenarios.
As a promising unsupervised learning paradigm ,they have led to great empirical success in computer vision tasks and gain the advanced performance in deep representational learning with unsupervised contrastive pre-training \cite{grill2020bootstrap, li2020prototypical}. Contrastive methods trained on unlabelled ImageNet data and evaluated with a linear classifier now surpass the accuracy of supervised AlexNet \cite{henaff2020data}. And Contrastive pre-training on ImageNet successfully transfers to other downstream tasks and outperforms the supervised pre-training counterparts \cite{he2020momentum}.

Many studies focus on constructing soft labels between data pairs under unsupervised settings artificially using the following two strategies.
One is to use clustering results as pseudo-labels to guide pair construction \cite{caron2018deep, asano2019self}. The objective in clustering is tractable, but it does not scale well with the dataset as it requires a pass over the entire dataset to form image codes (i.e., cluster assignments) that are used as targets during training.
Another, more direct and common approach is to treat each instance as a class and construct data pairs through data augmenting \cite{dosovitskiy2015discriminative, li2020contrastive, misra2020self, jing2020self}. Specifically, positive pairs consist of two enhanced views of the same instance, and other pairs are defined as negative pairs \cite{bojanowski2017unsupervised}.

The difference between the proposed AHCL method and diverse previous works like \cite{ge2020self,kim2020adversarial} represents that soft labels were pre-given for contrastive learning, while our method can learn the corresponding labels through deep features adaptively. Accordingly, our model constructs the deep affinity relationship  via the adaptive labels with the embedding of samples simultaneously. Therefore, it can be effectively used for spectral clustering instead of generating the Laplacian laboriously \cite{von2007tutorial,liu2018spectral,zhang2018understanding}.
\section{Framework of Adaptive Homotopy learning}
\label{Ch3}

In the paper, a framework regarding the adaptive homotopy learning (AH) is proposed, which can be applied to various homotopy-based models for directly adaptive learning. Motivated by integration between the homotopy idea as Eq. (\ref{eq:H2}) and embedding extraction methods, we propose an original homotopy model regarding the feature extraction as
\begin{equation}
      \min_{\bm{z}} \quad y f(\bm{z} \mid \bm{x})+(1-y) g(\bm{z} \mid \bm{x}), \quad {y} \in[0,1],
      \label{eq:HM0}
\end{equation}
where $y$ is a ratio coefficient and $y=1$ means that the prior feature $\bm{x}$ is completely mapped to the target space $\bm{D}$ through the observation function $f(\bm{z} \mid \bm{x})$.

However, like most homotopy models, the model relies heavily on prior conditions to determine the value of the coefficient $y$, so they are unable to implement adaptive learning of $y$ effectively. It is due to the fact that the model Eq. (\ref{eq:HM0}) is linear with respect to weight $y$, such that the model cannot complete the adaptive updating of weight $y$ by taking the extreme value condition with respect to $y$. Thus, the model (\ref{eq:HM0}) cannot perform the adaptive
learning automatically.

To solve the problem that the model cannot be updated adaptively, we begin with the above contradiction points, focus on optimizing the expression of the ratio coefficient $y$, and then put forward a new adaptive homotopy model (updating mechanism). Motivated by Maclaurin series as
\begin{equation}
      \ln (1-y)=-\sum_{n=1}^{\infty} \frac{y^{n}}{n}=-y-\frac{y^{2}}{2}-\frac{y^{3}}{3}-\cdots-\frac{y^{n}}{n}-\cdots \quad , \forall y \in(0,1),
      \label{eq:Maclaurin}
\end{equation}
we have the equivalent infinitesimal for $y \rightarrow  0$ as
\begin{equation}
      y \sim -\ln (1-y) \quad , \        1-y \sim  -\ln y \quad.
      \label{eq:infinitesimal}
\end{equation}

Therefore, Maclaurin approximation regarding the weight $y$ is positively correlated and equivalent. Based on the duality and Maclaurin approximation whose coefficient on weight $y$ is nonlinear and differentiable, we propose an adaptive homotopy model defined as
\begin{equation}
      \min _{\bm{z} , 0 \leq y\leq 1 } \quad -\ln (1-y)  f(\bm{z} \mid \bm{x})-\ln (y)  g(\bm{z} \mid \bm{x}).
      \label{eq:ACL}
\end{equation}
Since the Eq. (\ref{eq:ACL}) is able to determine the extremum by taking the derivative of the weight $y$, the model can be updated adaptively. Furthermore, we can deduce the adaptive weight (update rule) for $y$ as
\begin{equation}
      y=\frac{g}{f+g}.
      \label{eq:y}
\end{equation}
It is worth noting that when  $f=0 \ or\ g=0 $, Eq. (\ref{eq:ACL}) is trivial due to the fact that the domain of the $\ln$ function cannot be $0$. To handle the problem, we further optimize the update strategy of the adaptive weight $y$ by adopting the piecewise principle as
\begin{equation}
      y=\left\{\begin{array}{cc}
            0,             & g \rightarrow 0           \\
            \specialrule{0em}{0.5ex}{0.5ex}
            \frac{g}{f+g}, & f \neq 0 \ \& \  g \neq 0 \\
            \specialrule{0em}{0.5ex}{0.5ex}
            1,             & f \rightarrow 0.
      \end{array}\right.
      \label{eq:Y}
\end{equation}

Nevertheless, if Eq. (\ref{eq:ACL}) is updated by Eq. (\ref{eq:Y}), its loss could result in an explosion or collapse. For example, the case of $\infty \cdot 0$ may happen when $ \  g_{i j}\rightarrow 0 $.
To avoid the referred problem, our objective function is decoupled into
\begin{equation}
      \min_{\bm{z} , 0 \leq y\leq 1} \quad y f(\bm{z} \mid \bm{x})+(1-y) g(\bm{z} \mid \bm{x}), \quad \mathrm{y} \in[0,1],
      \label{eq:AH}
\end{equation}
since Eqs. (\ref{eq:HM0}) and (\ref{eq:ACL}) are homotopy equivalent. Besides that, the adaptive weight $y$ is still updated by Eq. (\ref{eq:Y}). In sum, the adaptive homotopy model can be optimized via Algorithm \ref{algo:algorithm1}.
\IncMargin{1em}
\begin{algorithm}
      \caption{Algorithm for adaptive homotopy model}
      \label{algo:algorithm1}
      \KwIn{A matrix:$\bm{X}_{n \times m} =\left[\bm{x}_{1}, \bm{x}_{2}, \bm{x}_{3}, \ldots, \bm{x}_{n}\right]^{T} $}
      \KwOut{The embedding:$\bm{Z}_{n \times k} =\left[\bm{z}_{1}, \bm{z}_{2}, \bm{z}_{3}, \ldots, \bm{z}_{n}\right]^{T} $ and a corresponding coefficient vector: $\bm{y}_{n}$ }
      \While{not converge}
      {
            $\textbf{extracting:}$ {the embedding $\bm{z_{i}}\leftarrow\bm{x}_{i}$}\;
            \For{$\textit{i}=1;\textit{i} \le \textit{n};\textit{i++}$}
            {
                  $\textbf{compute:}${the positive objective function: $f(\bm{z}_i \mid \bm{x}_{i})$ and the negative objective function:$g(\bm{z}_i \mid \bm{x}_{i})$ } \;
                  $\textbf{update:} $ {adaptive soft label: $y_{i}$ by Eq. (\ref{eq:Y})} \;
            }
            $\textbf{update:}$ {$\bm{z}, y$ with Eq. (\ref{eq:Y},\ref{eq:AH}) through gradient descent}\;
      }
\end{algorithm}
\DecMargin{1em}

\section{Deep Graph Representation}
\label{Ch4}

In order to extract better deep features adaptively, we introduce the proposed AH into the supervised contrastive learning model. Therefore, an adaptive and unsupervised contrastive learning model is proposed as
\begin{equation}
      \min_{\bm{z} , 0 \leq y\leq 1}  \quad \frac{1}{2 n} \sum_{i=1}^{n}  \sum_{j=1}^{n} y_{i j}  f(\bm{z}_{i} \mid \bm{x}_{i},\bm{z}_{j} \mid \bm{x}_{j})+\left( 1-y_{i j} \right)  g(\bm{z}_{i} \mid \bm{x}_{i},\bm{z}_{j} \mid \bm{x}_{j}),
      \label{eq:CL1}
\end{equation}
where label $ y_{i j}\  \epsilon  \ [0,1] $ denotes the similarity between the $i$-th data point $x_i$ and the $j$-th data point $x_j$. In other words, bigger $y$ denotes a more similar data pair.

For brevity, $f(\bm{z}_{i} \mid \bm{x}_{i},\bm{z}_{j} \mid \bm{x}_{j})$ and $g(\bm{z}_{i} \mid \bm{x}_{i},\bm{z}_{j} \mid \bm{x}_{j})$ are simplified as $f(\bm{z}_i,\bm{z}_j)$ and $g(\bm{z}_i,\bm{z}_j)$ respectively, where both functions are used to measure and fit the degree of affinity between deep features of data pairs in feature space $\bm{Z}$. Specifically speaking, $f(\bm{z}_i,\bm{z}_j)$ is used to shrink the distance of similar feature pairs in the feature space, so that they can aggregate as much as possible. On the contrary, $g(\bm{z}_i,\bm{z}_j)$ is used to amplify the distance of dissimilar feature pairs, so that they can be clearly separated as much as possible. To simplify the complexity of our proposed model, Euclidean distance is utilized as the affinity assessment in our work. In general, Euclidean distance between $\bm{z}_{i}=\left(z_{1}, z_{2}, z_{3}, \ldots, z_{n}\right)^{T}$ and $\bm{z}_{j}=\left(z_{1}, z_{2}, z_{3}, \ldots, z_{n}\right)^{T}$ is denoted as
\begin{equation}
      d_{i j}= \|\bm{z}_{i}-\bm{z}_{j} \Vert_2 .
      \label{eq:dist}
\end{equation}
From the perspective of contrastive learning, we define specific expressions of $ \bm{f}$ and $\bm{g}$ respectively to extract the embedding as
\begin{equation}
      \begin{array}{l}
            f_{i j}=d_{i j}^{2} \\
            g_{i j}=\max \left(m-d_{i j}, 0\right)^{2},
      \end{array}
      \label{eq:fg}
\end{equation}
where $m>0$ is a margin. $\bm{z}_i$ and $\bm{z}_j$ is dissimilar when the $d_{ij}$ is larger than $m$. Therefore, we have the update strategy regarding the adaptive label $y_{i j}$ according to Eq. (\ref{eq:Y}), which can be defined as
\begin{equation}
      y_{i j}=\left\{
      \begin{array}{lr}
            0                               & , d_{i j} \geq m
            \\
            \specialrule{0em}{0.5ex}{0.5ex}
            \frac{g_{i j}}{f_{i j}+g_{i j}} & , 0<d_{i j}<m
            \\
            \specialrule{0em}{0.5ex}{0.5ex}
            1                               & , d_{i j}=0 .
            \\
      \end{array}\right.
      \label{eq:y2}
\end{equation}
where $y$ is normalized to  $[0, 1]$,  $0$  means no correlation, and $1$ means perfect correlation. Accordingly, from Eqs. (\ref{eq:CL1}), (\ref{eq:dist}), and (\ref{eq:y2}), we propose the adaptive contrastive learning model as
\begin{equation}
      \mathcal{L}_{AHCL} = \frac{1}{2 n} \sum_{i=1}^{n}  \sum_{j=1}^{n} y_{i j}  d_{i j}^{2}+\left(1-y_{i j}\right)  \max \left(m-d_{i j}, 0\right)^{2}.
      \label{eq:ACL2}
\end{equation}

Since the model obtained above is absolutely unsupervised, its performance is largely affected by the deep feature $\bm{Z}$ extracted by the encoder. To ensure that the learned embedding $\bm{Z}$ is always closely related to the corresponding input data $\bm{X}$, we utilize a decoder to reconstruct the original $\bm{X}$ to fine-tune and optimize the encoder according to the reconstruction error.

The fitting error for the decoder is measured by MSE functions as
\begin{equation}
      \mathcal{L}_{MSE} = \frac{1}{n} \sum_{i=1}^{n} \sum_{j=1}^{m}\left(x_{i j}-\hat{x}_{i j}\right)^{2},
      \label{eq:MSE}
\end{equation}
where $\bm{x}$ is the input vector and $\bm{\hat{x}}$ is the corresponding reconstructed vector, $n$ is the total number of vectors, and $m$ is the dimension of vectors.

To improve the performance of feature extraction,  we devise the total objective function via a linear combination of Eqs. (\ref{eq:ACL2}) and (\ref{eq:MSE}) as
\begin{equation}
      \mathcal{L} =\mathcal{L}_{AHCL}+\gamma  \mathcal{L}_{MSE}, \gamma \in \mathbb{R}^{+},
      \label{eq:TL}
\end{equation}
where $\gamma > 0$ is a hyper-parameter.
To sum up, AHCL can be optimized via Algorithm \ref{algo:algorithm2} to learn the deep graph representation.
\IncMargin{1em}

\begin{algorithm}[]
      \caption{Algorithm for learning the deep graph representation}
      \label{algo:algorithm2}
      \linespread{1.15}
      \KwIn{An image matrix: $\bm{X}_{n \times m} =\left[\bm{x}_{1}, \bm{x}_{2}, \bm{x}_{3}, \ldots, \bm{x}_{n}\right]^{T} $}
      \KwOut{An affinity matrix: $\bm{Y}_{n \times n}$ and embedding: $\bm{Z}_{n \times k}$}
      $\textbf{initialize:}${ The distance margin: $m \in (0,1]$ ,and the loss ratio:$\gamma \in (0,1]$\;}
      \tcp*[h]{Training}\;
      \While{not converge}
      {
            $\textbf{extracting:}$ {the embedding $\bm{Z}_{n \times k} \leftarrow \bm{X}_{n \times m}$ through Algorithm. \ref{algo:algorithm1}}\;
            $\textbf{reconstruction:}$ {$\bm{\hat{X}}_{n \times n}\leftarrow\bm{Z}_{n \times k}$ by $\bm{Decoder}$}\;
            $\textbf{update:}$      {$\bm{Encoder} ,\bm{Decoder}$ through gradient descent by Eq. (\ref{eq:ACL2}, \ref{eq:MSE}, \ref{eq:TL})} \;
      }
      \tcp*[h]{Generating Deep Graph}\;
      $\textbf{generate:}$    {Deep graph representation $\bm{Y}_{n \times n} \leftarrow \bm{Z}_{n \times k}$  through Eq. (\ref{eq:dist},\ref{eq:fg},\ref{eq:y2})}\;
\end{algorithm}
%  $$\sqrt{
%       \left(\begin{array}{cccc}
%       \left\|Z_{1}\right\|^{2} & \left\|Z_{1}\right\|^{2} & \cdots & \left\|Z_{1}\right\|^{2} \\
%       \left\|Z_{2}\right\|^{2} & \left\|Z_{2}\right\|^{2} & \cdots & \left\|Z_{2}\right\|^{2} \\
%       \vdots & \vdots & \ddots & \vdots \\
%       \left\|Z_{n}\right\|^{2} & \left\|Z_{n}\right\|^{2} & \cdots & \left\|Z_{n}\right\|^{2}
%       \end{array}\right)+\left(\begin{array}{cccc}
%       \left\|Z_{1}\right\|^{2} & \left\|Z_{2}\right\|^{2} & \cdots & \left\|Z_{n}\right\|^{2} \\
%       \left\|Z_{1}\right\|^{2} & \left\|Z_{2}\right\|^{2} & \cdots & \left\|Z_{n}\right\|^{2} \\
%       \vdots & \vdots & \ddots & \vdots \\
%       \left\|Z_{1}\right\|^{2} & \left\|Z_{2}\right\|^{2} & \cdots & \left\|Z_{n}\right\|^{2}
%       \end{array}\right)-2 \times \bm{Z} \bm{Z}^{T}
%                         }$$
\section{Experiment}
\label{Ch5}
In this section, we evaluated the effectiveness and performance of the proposed AHCL on five unsupervised benchmark datasets by analyzing the experiments. Firstly, we verified the superiority of AHCL by comparing to six clustering frameworks. To avoid the performance impact of our backbone network, we verified the significance of AHCL by clustering on original features directly and deep embedding extracted using only MSE respectively. Finally, the convergence of AHCL and the theoretical analysis mentioned above are verified including visualization.
\subsection{Datasets}
To show the superiority of our method AHCL framework, it is evaluated on five benchmark datasets. They are four different types  of image datasets (UMIST \cite{UMIST}, COIL20 \cite{COIL20}, USPS \cite{USPS}, and FASHION-MNIST-test \cite{FASHION}) and a UCI \cite{UCI} dataset (PALM). For simplicity , FASHION-MNIST-test is denoted by FASHION. A brief description of these datasets is summarized in Table \ref{tab:dataset-table}, including the number of data samples, the number of features contained in each sample, the number of classes, and the type of each dataset.

\subsection{Evaluation metrics}
In unsupervised feature representation, the learned embedding is usually evaluated by clustering. Hence, two widely-used standard clustering evaluation metrics are used as Accuracy (ACC) and Normalized Mutual Information (NMI) to evaluate our method \cite{amigo2009comparison,NMI}.
Concisely, the greater values of ACC and NMI, the better the clustering effect, which also means the better the performance of the deep graph representation. The results of NMI are not changed by the arrangement of clusters (classes).
% They are normalized to  $[0, 1]$, where $0$  means no correlation and $1$ means perfect correlation. 
For completeness, we define ACC as
\begin{figure}[htp]
      \centering
      \subfigure[epoch.1 (NMI = 65.17\%)]{
            \begin{minipage}[t]{0.33\linewidth}
                  \centering
                  \includegraphics[width=1.7in, height=1.7in]{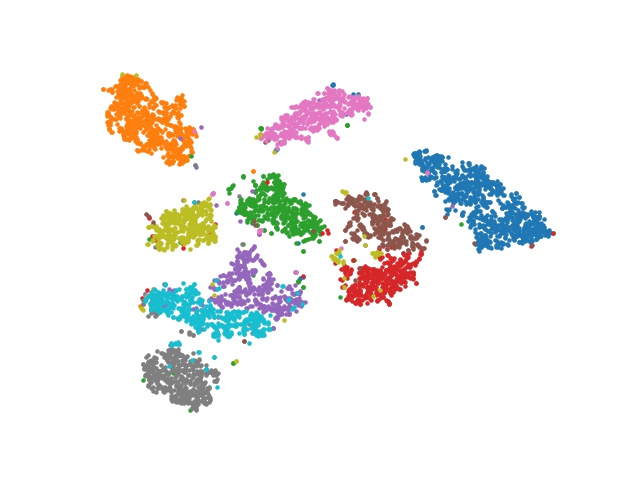}
                  %\caption{fig1}
            \end{minipage}%
      }%
      \subfigure[epoch.20 (NMI = 79.17\%)]{
            \begin{minipage}[t]{0.33\linewidth}
                  \centering
                  \includegraphics[width=1.7in, height=1.7in]{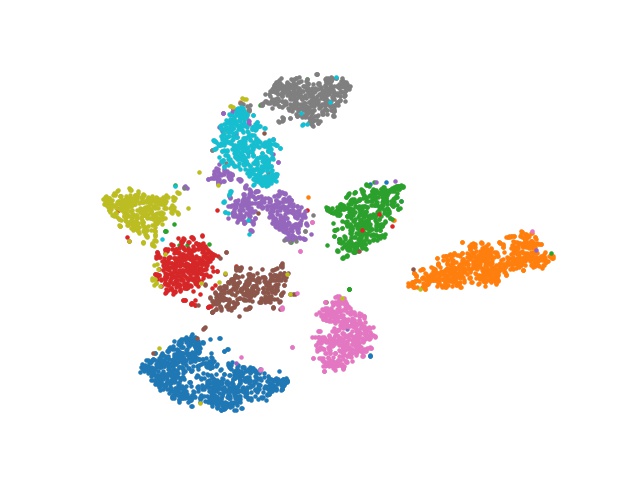}
                  %\caption{fig1}
            \end{minipage}%
      }%
      \subfigure[epoch.140 (NMI = 82.46\%)]{
            \begin{minipage}[t]{0.33\linewidth}
                  \centering
                  \includegraphics[width=1.7in, height=1.7in]{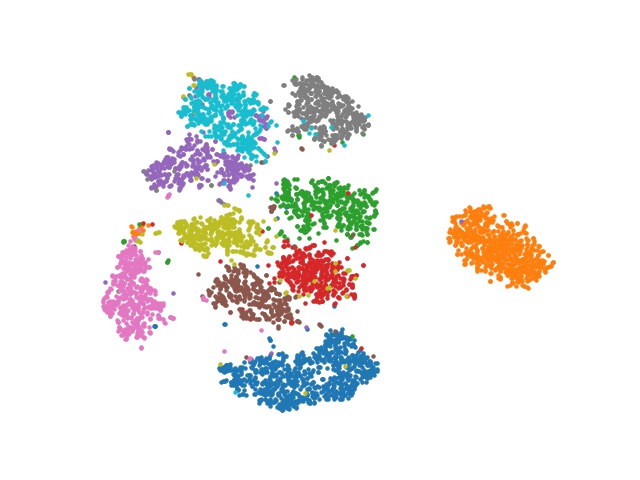}
                  %\caption{fig1}
                  %\caption{fig2}
            \end{minipage}
      }%
      \centering
      \caption{The evolution of deep instance features in plane space by T-SNE mapping during training on USPS. The same color corresponds to belong to the same class.}
      \label{Fig:2}
\end{figure}
\begin{table}[H]
      \centering
      \linespread{1.25}
      \caption{A summary of the benchmark datasets used for evaluations}
      \label{tab:dataset-table}
      \begin{tabular}{@{}ccccc@{}}
            \toprule
            Dataset & Samples & Dimensions & Classes & Type          \\
            \midrule
            UMIST   & 575     & 1024       & 20      & Face image    \\
            COIL20  & 1440    & 1024       & 20      & Object image  \\
            PALM    & 2000    & 256        & 100     & UCI           \\
            USPS    & 9298    & 256        & 10      & Digital image \\
            FASHION & 10000   & 784        & 10      & Object image  \\
            \bottomrule
      \end{tabular}
\end{table}
\begin{equation}
      ACC=\max _{m} \frac{\sum_{i=1}^{n} \mathbf{1}\left\{l_{i}=m\left(c_{i}\right)\right\}}{n},
\end{equation}
where $l_i$ and $c_i$ are the ground-truth label and predicted cluster label of data point $x_i$, respectively.
NMI calculates the normalized measure of similarity between two labels of the same data as
\begin{equation}
      NMI=\frac{I(l ; c)}{\max \{H(l), H(c)\}},
\end{equation}
where $I(l; c)$ denotes the mutual information between true label $l$ and predicted cluster $c$, and $\mathrm{H}$ represents their entropy.

Note that all the experimental data are the average results of running the corresponding methods $10$ times, and the error range is no more than $2\%$..

\subsection{Experimental Setup}
In our experiments, a Siamese network with shared weights \cite{hadsell2006dimensionality} is served as the encoder, which is consisted of three sets of residual blocks with a set of convolution. The decoder consists of three transposed convolutional layer. The dimensions of the embedding $\bm{Z}$ is set to $512$ or $128$ through two fully connected layers. Except for the last layer of Decoder being activated by Sigmoid, the other layers adopt ReLU as the non-linear activation. Moreover, structure of AHCL is shown in Figure \ref{Fig:1}.

As for two hyper-parameter, we set $m \rightarrow 0.7 \sim 0.8$ and $\gamma \rightarrow  0.01 \sim  0.001$. We use Adaptive Moment Estimation (Adam) as the optimizer to compute adaptive learning rates for each parameter. In addition, the learning rate is set to decrease with the step size, and our initial learning rate is $1e-3$.
For data loading, the training data pairs of each epoch adopt the strategy of random cross combination and expansion, which enables the network to better aggregate similar feature Spaces and separate dissimilar feature Spaces meanwhile. For instance, FASHION-MNIST-Test (simplified FASHION) contains a total of 10,000 testing samples. We randomly selected $2000$ samples from the total samples as a batch at each iteration and set a total of $20$ such iterations for each epoch.
Note that codes of all the experiments are implemented under the PyTorch-1.4.0 and python-3.7.9 on an Ubuntu 18.04.2 LTS server with an NVIDIA GeForce GTX 1080Ti GPU.
\begin{table}[h]
      \centering
      \caption{ The clustering performance  (\%) of ten methods on five challenging benchmarks datasets}
      \label{tab:table2}
      \resizebox{1\textwidth}{!}{%
            \linespread{1.25}
            \begin{tabular}{@{}ccccccccccc@{}}
                  \toprule
                  Dataset                     &
                  \multicolumn{2}{c}{UMIST}   &
                  \multicolumn{2}{c}{COIL20}  &
                  \multicolumn{2}{c}{PALM}    &
                  \multicolumn{2}{c}{FASHION} &
                  \multicolumn{2}{c}{USPS}                                                                                                \\ \midrule
                  Metrics                     & ACC               & NMI   & ACC   & NMI   & ACC   & NMI   & ACC   & NMI   & ACC   & NMI   \\ \midrule
                  K-Means                     & 42.87             & 65.47 & 58.26 & 74.58 & 70.39 & 89.98 & 47.65 & 51.09 & 64.83 & 62.67 \\
                  SC-Ncut                     & 60.05             & 77.54 & 66.74 & 83.00 & 61.19 & 85.21 & 50.90 & 51.81 & 67.85 & 76.25 \\

                  CAN                         & 69.62             & 87.75 & 84.10 & 90.93 & 88.10 & 97.08 & -     & -     & 67.96 & 78.85 \\
                  DEC                         & 36.47             & 56.96 & 74.35 & 90.37 & 27.45 & 55.22 & 51.77 & 54.68 & 42.30 & 48.71 \\
                  DFKM                        & 45.47             & 67.04 & 60.21 & 76.81 & 67.45 & 86.74 & 57.12 & 60.45 & 73.42 & 71.58 \\
                  AE                          & 62.61             & 82.66 & 73.38 & 81.81 & 85.71 & 92.11 & 56.95 & 59.23 & 70.70 & 70.05 \\
                  VAE                         & 54.61             & 73.54 & 75.69 & 85.38 & 86.88 & 94.96 & 59.10 & 56.23 & 71.20 & 72.03 \\
                  GAE                         & 61.91             & 80.24 & 69.10 & 86.45 & 88.45 & 94.87 & -     & -     & 76.63 & 76.02 \\
                  AdaGAE                      &
                  \underline{83.48}           & \underline{91.03} &
                  \textbf{93.75}              & \textbf{98.36}    &
                  \underline{95.25}           & \underline{98.18} &
                  \underline{62.67}           & \underline{65.79} &
                  \textbf{91.96}              & \textbf{84.81}                                                                            \\
                  SpectralNet                 & 52.53             & 79.99 & 75.69 & 88.93 & 76.94 & 92.85 & -     & -     & 70.98 & 74.83 \\
                  \textbf{AHCL(ours)}         &
                  \textbf{90.78}              &
                  \textbf{95.47}              &
                  \underline{90.14}           & \underline{97.02} &
                  \textbf{96.45}              &
                  \textbf{99.32}              &
                  \textbf{65.40}              &
                  \textbf{70.48}              &
                  \underline{89.33}           & \underline{84.07}                                                                         \\ \bottomrule
            \end{tabular}
      }
\end{table}
\begin{figure}[htp]
      \centering
      \subfigure[epoch.1 (ACC = 65.17\%)]{
            \begin{minipage}[t]{0.33\linewidth}
                  \centering
                  \includegraphics[width=1.55in, height=1.6in]{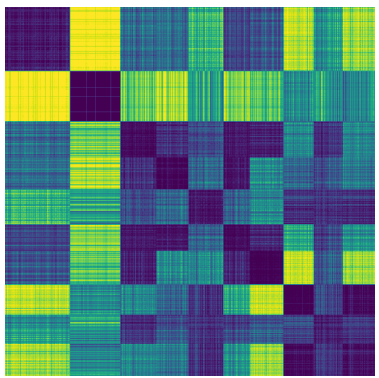}\\
                  % \vspace{0.02cm}
                  %\caption{fig1}
            \end{minipage}%
      }%
      \subfigure[epoch.20 (ACC = 68.16\%)]{
            \begin{minipage}[t]{0.33\linewidth}
                  \centering
                  \includegraphics[width=1.55in, height=1.6in]{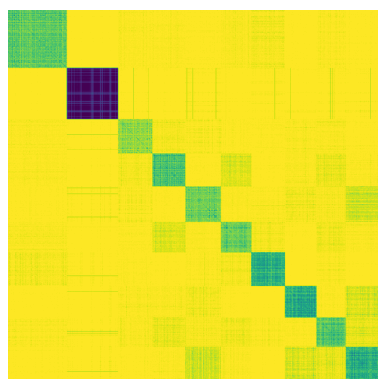}
                  %\caption{fig1}
                  %\caption{fig2}
            \end{minipage}%
      }%
      % \subfigure[epoch.60 (ACC = 77.19)]{
      %       \begin{minipage}[t]{0.33\linewidth}
      %             \centering
      %             \includegraphics[width=1.0in]{./graph/160.png}
      %             %\caption{fig1}
      %             %\caption{fig2}
      %       \end{minipage}
      % }%
      \subfigure[epoch.140 (ACC = 87.52\%)]{
            \begin{minipage}[t]{0.33\linewidth}
                  \centering
                  \includegraphics[width=1.55in, height=1.6in]{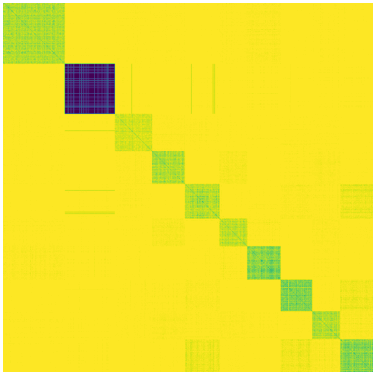}
                  %\caption{fig1}
                  %\caption{fig2}
            \end{minipage}
      }%
      \centering
      \caption{The evolution of deep affinity relationship during training on USPS. The color piece between the corresponding categories is clearly distinguished.}
      \label{Fig:3}
\end{figure}
\subsection{Results and Analysis}
\subsubsection{Comparisons with baselines of clustering}
To evaluate the performance of AHCL, ten representative state-of-the-art clustering methods serve as competitors. To ensure fairness, 3 clustering methods without neural networks are used, including K-Means \cite{zhang2019deep}, Spectral Clustering with Normalized Cut(SC-Ncut) \cite{SC-Ncut} and CAN \cite{CAN}. Five deep clustering methods for general data, including DEC \cite{DEC}, DFKM \cite{DFKM}, AE \cite{AE}, VAE \cite{VAE} and SpectralNet \cite{SpectralNet}, also serve as an important baseline. Furthermore, two GNN-based method, GAE \cite{GAE} and AdaGAE \cite{AdaGAE}, are also used. All codes are downloaded from the home pages of authors. The concrete information of their settings can be found in supplementary. In terms of compared methods, when the released code is not publicly available, or running the released code is not practical, we put dash marks (-) instead of the corresponding results.

All clustering results of comparison methods and our AHCL about ACC and NMI are shown in the Table \ref{tab:table2}. The best results of both competitors and AHCL are highlighted in boldface, while the second-best results are underlined. From Table \ref{tab:table2}, we we conclude that: our method performs significantly better on five benchmark datasets than nine compared baselines methods.

\begin{table}[t]
      \centering
      \caption{Clustering performances (\%) of Ablation modules on five datasets }
      \label{tab:table3}
      \resizebox{1\textwidth}{!}{%
            \begin{tabular}{@{}cccccccccccc@{}}
                  \toprule
                  \linespread{1.25}
                                              &
                  Dataset                     &
                  \multicolumn{2}{c}{UMIST}   &
                  \multicolumn{2}{c}{COIL20}  &
                  \multicolumn{2}{c}{PALM}    &
                  \multicolumn{2}{c}{FASHION} &
                  \multicolumn{2}{c}{USPS}                                                                                              \\ \midrule
                                              & Metrics & ACC   & NMI   & ACC   & NMI   & ACC   & NMI   & ACC   & NMI   & ACC   & NMI   \\ \midrule
                                              & KM-Z    & 43.30 & 65.67 & 58.26 & 74.58 & 70.39 & 89.98 & 47.65 & 51.09 & 64.83 & 62.67 \\
                  Surface                     & SC-Z    & 50.15 & 74.54 & 66.74 & 83.00 & 61.19 & 85.21 & 50.90 & 51.81 & 67.85 & 76.25 \\
                                              & SC-Y    & 34.79 & 59.49 & 55.01 & 72.84 & 55.97 & 83.97 & 43.61 & 47.73 & 53.23 & 52.02 \\ \midrule
                                              & KM-Z    & 49.91 & 75.14 & 80.72 & 81.81 & 74.96 & 89.22 & 59.10 & 56.23 & 68.33 & 66.51 \\
                  MSE                         & SC-Z    & 62.61 &
                  \underline{82.66}           & 75.69
                                              & 85.38   & 86.88 & 94.96 & 52.42 & 50.91 & 77.72 & 76.84                                 \\
                                              & SC-Y    & 42.26 & 69.37 & 67.43 & 78.68 & 85.35 & 95.23 & 54.94 & 55.88 & 71.20 & 72.03 \\ \midrule
                  \textbf{}                   & KM-Z
                                              &
                  \underline{64.87}           & 68.18   & 78.26 & 84.88 & 92.95 & 98.05 &
                  \underline{71.15}           &
                  \underline{65.13}           & 70.13   & 66.98                                                                         \\
                  \textbf{AHCL}               &
                  SC-Z                        & 60.00   & 77.51 &
                  \underline{81.60}           &
                  \underline{90.71}           &
                  \textbf{96.45}              &
                  \textbf{99.32}              & 64.52   & 70.90 &
                  \underline{85.48}           &
                  \underline{83.51}                                                                                                     \\
                  \textbf{}                   &
                  SC-Y                        &
                  \textbf{90.78}              &
                  \textbf{95.47}              &
                  \textbf{90.14}              &
                  \textbf{97.02}              &
                  \underline{94.30}           &
                  \underline{99.05}           &
                  \textbf{65.14}              &
                  \textbf{71.80}              &
                  \textbf{89.33}              &
                  \textbf{84.07}                                                                                                        \\ \bottomrule
            \end{tabular}
      }
\end{table}

\begin{figure}[h]
      \centering
      \begin{tabular}{m{0.165\textwidth}m{0.165\textwidth}m{0.165\textwidth}m{0.165\textwidth}m{0.2\textwidth}}
            \subfigure[UMIST]{
                  \centering
                  \includegraphics[width=1.1in, height=1.1in]{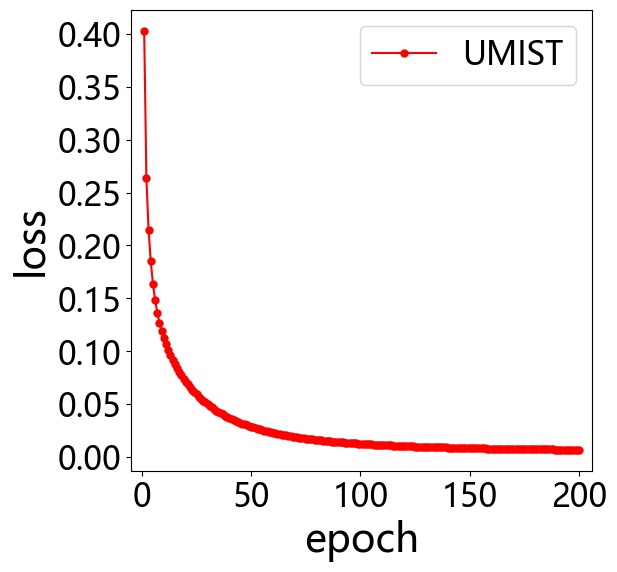}
                  %\caption{fig1}
            }
             & \subfigure[COIL20]{
                  \centering
                  \includegraphics[width=1.1in, height=1.1in]{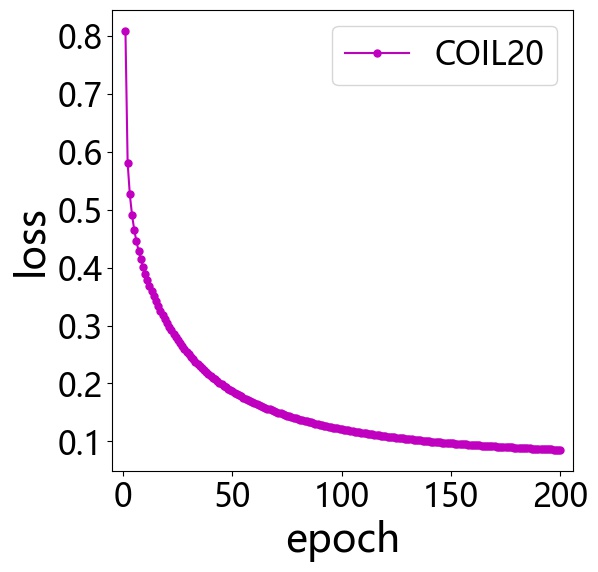}
                  %\caption{fig1}
            }
             & \subfigure[PALM]{
                  \centering
                  \includegraphics[width=1.1in, height=1.1in]{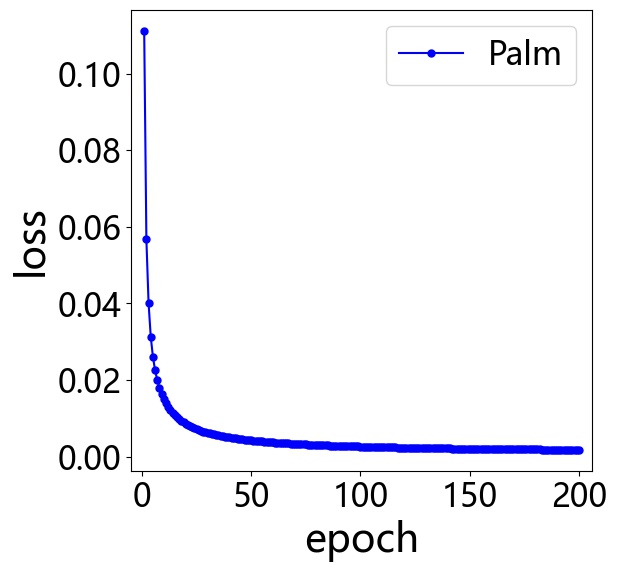}
                  %\caption{fig1}
            }
             & \subfigure[FASHION]{
                  \centering
                  \includegraphics[width=1.1in, height=1.1in]{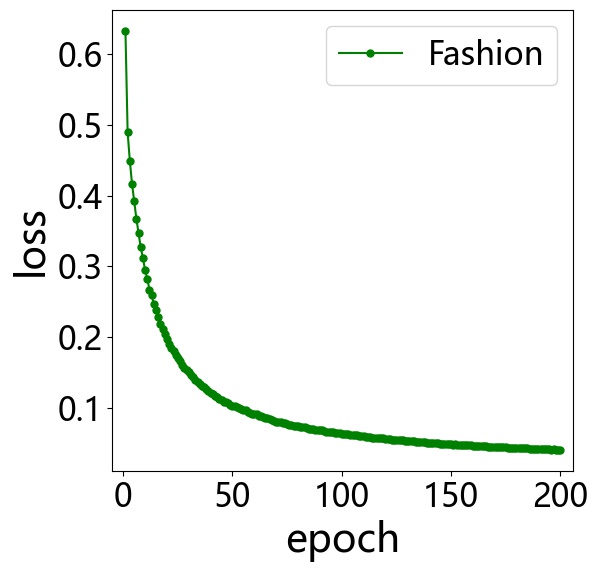}
                  %\caption{fig1}
            }
             & \subfigure[USPS]{
                  \centering
                  % \flushleft
                  \includegraphics[width=1.1in, height=1.1in]{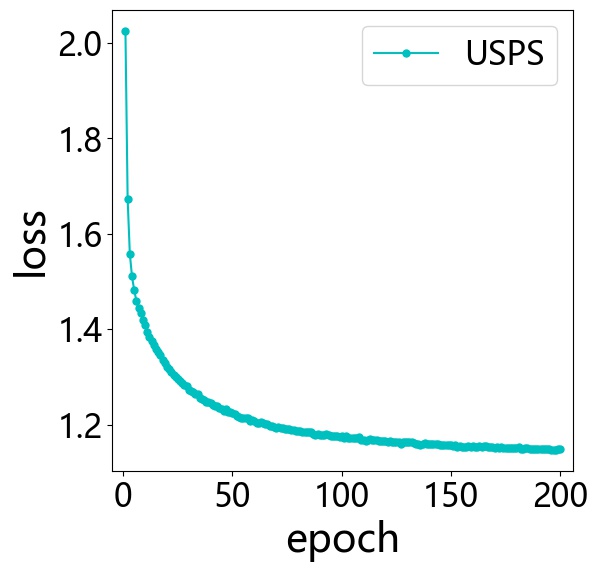}
                  %\caption{fig1} 
            }
      \end{tabular}
      \caption{Convergence curve of AHCL on five datasets}
      \label{Fig:loss}
\end{figure}

\subsubsection{Ablation Study}
To examine how much our model relies on the structure of the backbone network, we set up three groups of ablation experiments on the surface features, the deep features learned from our same Siam-AE structure but using MSE as the loss function only and the embedding extracted by the AHCL respectively. For each ablation experiment, they are as follows: direct k-means clustering on feature $\bm{Z}$ (denoted as KM-Z), direct spectral clustering on feature $\bm{Z}$ (SC-Z) , and spectral clustering on soft label matrix $\bm{Y}$ obtained from feature $\bm{Z}$ as the affinity matrix (SC-Y). The Clustering results are reported in Table \ref {tab:table3}. The results suggest that the backbone network contributes to the clustering performance. What is more obvious is that AHCL can give the network higher-performance of presentation learning. Therefore, the deep graph representation learned via the AHCL $\bm{Y}$ is reasonable.

To understand how AHCL works intuitively, we visualize the evolution process with the training epoch from distribution of both the instance-level embedding $\bm{Z}$ and cluster-level blocks of the affinity representation matrix $\bm{Y}$.  As shown in Figure \ref{Fig:2}, the two-dimensional mapping of deep features by T-SNE become more and more convergent when the corresponding data points are same class. Meanwhile, it is more and more separate for the spatial distribution that don't belong to the same class. Besides, Figure \ref{Fig:3} illustrates the evolving of learned affinity relationship between classes constantly, in which the pieces of the same kind become more and more obvious and pure.

\subsubsection{Convergence analysis}
The convergence of our model is shown in the Figure \ref{Fig:loss}. The vertical axis represents the objective function value, and the horizontal axis represents the number of iterations. It can be observed that the rapid convergence is performed for all the five benchmark datasets. In particular, our model reaches steady convergence when epoch is more than 60. Furthermore, AHCL behaves consistently on all datasets due to the adaptive mechanism of affinity weights.

\section{Conclusion}

In this paper, we propose a novel adaptive homotopy framework (AH). Compared with the classic homotopy model, it can adaptively obtain the optimal parameters via the Maclaurin duality. Particularly, aiming to extend the contrastive learning from weak-supervised learning to unsupervised learning, we apply the proposed adaptive homotopy framework to contrastive learning (AHCL). Thus, AHCL not only can  adaptively learn the soft labels but also extract the deep features without any prior information. Furthermore, based on the affinity matrix formulated by the adaptively learned soft labels, a novel deep Laplacian graph can be constructed to explore and fit the topology of deep features. On the benchmark datasets, the proposed AHCL achieves better results.

\newpage
{
      \bibliography{./citations}
      \bibliographystyle{IEEEtran}
}

\end{document}